\pdfoutput=1
\documentclass{svmult}
\begin{document}

\title*{A Distributed Process Infrastructure for a Distributed Data Structure}
\author{Marko A. Rodriguez}

\institute{
	T-7, Center for Non-Linear Studies \\ 
	Los Alamos National Laboratory \\
	\texttt{marko@lanl.gov}
}

\maketitle

\begin{abstract}
The Resource Description Framework (RDF) is continuing to grow outside the bounds of its initial function as a metadata framework and into the domain of general-purpose data modeling. This expansion has been facilitated by the continued increase in the capacity and speed of RDF database repositories known as triple-stores. High-end RDF triple-stores can hold and process on the order of 10 billion triples. In an effort to provide a seamless integration of the data contained in RDF repositories, the Linked Data community is providing specifications for linking RDF data sets into a universal distributed graph that can be traversed by both man and machine. While the seamless integration of RDF data sets is important, at the scale of the data sets that currently exist and will ultimately grow to become, the ``download and index" philosophy of the World Wide Web will not so easily map over to the Semantic Web. This essay discusses the importance of adding a distributed RDF process infrastructure to the current distributed RDF data structure.
\end{abstract}

\section{Introduction}

The Semantic Web community has introduced a set of standards and protocols for representing, querying, and manipulating a distributed, directed labeled graph of web resources. The Uniform Resource Identifier (URI), as the foundation standard of the World Wide Web, provides the distributed address space for web resources. The Resource Description Framework (RDF), as the foundation standard of the Semantic Web, provides a data model for graphing web resources. In a local environment, URIs can be minted and related to one another in a relatively straightforward manner. What becomes difficult is making the link from the URIs of one environment to the URIs of another. That is, the same \textit{thing} can be given two unique URIs and the cross repository linking of such URIs is a difficult problem. With the World Wide Web, the author of an HTML document is responsible for the linking of web resources and this can be managed in a straightforward manner as HTML documents maintain a relatively small set of links. However, on the Semantic Web, there is no human consumable free-text \textit{per se}. Instead, every minutia of information is encoded as a statement (or triple). The amount of resources and links can grow fast and thus is only maintainable, for the most part, by machines. 

The Linked Data community expends much effort ensuring that RDF repositories are linked appropriately such that what emerges is a global web of data that is rich in content and able to be traversed by both man and machine \cite{berners:ldata2006}. However, traversing the Semantic Web is not quite the same as traversing the World Wide Web. For the human, it is reasonable to traverse from repository to repository exploring the Semantic Web in a manner similar to how the World Wide Web is traversed. However, for a machine, it is a different story. There will be orders of magnitude more resources and links on the Semantic Web. While a machine can crawl and pull the data to its local environment for processing, this becomes inefficient when the process' data requirements must span large parts of the Semantic Web graph. For distributed graph computations, a more efficient mechanism would be to migrate the process between RDF repositories so that information is not pulled back to the local environment, but instead, processed where the data is maintained. In other words, an efficient mechanism for processing the Semantic Web graph would be to move the process to the data, not the data to the process.

While the Semantic Web provides an infrastructure to support the distributed representation of a graph of resources, there currently does not exist a distributed process infrastructure for analyzing and manipulating this graph. For the World Wide Web, the search engine philosophy of ``download and index" has made it possible for end users to find information on the World Wide Web in a more efficient manner than by simply surfing and bookmarking. With modern commercial triple-stores scaling to the order of 10 billion triples, centralized indexing repositories will have to contend with not only the volume of data, but also the computational complexities of analyzing such a graph in a sophisticated way. The Semantic Web provides a much richer machine processable data structure than what is provided by HTML and thus, antiquated keyword search simply does not take significant advantage of what the Semantic Web is providing. Labeled graph query languages such as SPARQL are one method of retrieving data from the Semantic Web, but like keyword search, this is not the end story. The future of the Semantic Web will be rife with algorithms from many schools of thoughts. Many of these algorithms will compute across various repositories of the Semantic Web and will require a distributed Turing complete infrastructure to do so.

\section{The RDF Virtual Machine Model}

The notion of process migration for distributed computing is not new. For instance, the standards and technologies associated with Grid computing are aimed at facilitating the distributed and shared use of computing resources \cite{foster:grid2001}. Furthermore, the idea of a Semantic Grid has been proposed as an extension to the current Grid infrastructure to support the semantic annotation of computing resources to aid in discovery, reuse, etc. \cite{roure:semgrid2005}.
\begin{quote}
	``The Semantic Grid refers to an approach to Grid computing in which information, computing resources and services are described using the semantic data model."\footnote{Semantic Grid article is available at http://en.wikipedia.org/wiki/Semantic\_grid}
\end{quote}

The distributed process infrastructure discussed in this essay is aimed primarily at supporting the exploitation of the Semantic Web data structure directly as a data structure, not to augment an existing distributed processing infrastructure with the standards of the Semantic Web. The Web has provided a distributed address space that is infinite in size. It is possible to represent processes such they compute within this address space agnostic to the physical machines that support their execution. The idea of an RDF virtual machine (RVM) was introduced to facilitate this level of abstraction \cite{rodriguez:rvm2008}. With an RVM, the computing machine, the computing machine's software, and the data being processed are all represented in RDF and within the URI address space (as well as the blank and literal space). Some of the major consequences of the this computing model are
\begin{itemize}
	\item an RVM has no reference to the underlying physical (or virtual) machine executing it
	\item an RVM is constrained to the URI address space
	\item an RVM only executes RDF encoded software
	\item an RVM only processes RDF encoded data.
\end{itemize}
A primary tenet of the RVM model is that RDF is the lowest level of representation and in being the lowest level of representation, RDF is used to model all aspects of computing. Thus, RDF is used to define those data structures that are common in computing such as the lists, the sets, the arrays, the maps, the vectors, the stacks, the program counters, the heaps, the trees, the objects, the instructions---in general, the general-purpose modeling construct: the graph. Thus, RDF is used not just for representing metadata, RDF is used for representing data. When all information is placed in the URI address space, computing is at a new level of abstraction. This level of abstraction is a shared memory space that is built on the existing Web infrastructure.

As more data sets are linked to the growing Linked Data web, there will exist algorithms that will be burdened by the ``over the wire" speeds of the Internet. With the RVM model, the RVM can compute regardless of the physical machine that is supporting its execution. In this way, an RVM can traverse the Linked Data set not by pulling data to a local environment, but by actually moving between machines and more specifically, moving to those machines that are maintaining the subgraph of the Semantic Web that is of interest to the algorithm at particular points in time. 

For every architecture, there are drawbacks. In the RVM model of computing, the issues that are obvious are the issues of security, speed, and adoption.  With respect to security, the migration of process requires an infrastructure to protect the Semantic Web and the physical machines that support its representation. It is important to ensure that poorly or maliciously written RDF code does not destroy the integrity of an RDF data set, does not abuse the computational resources of a publicly available physical machine, and only accesses those aspects of an RDF data set that it has permission to access.

With respect to speed, the RVM model of distributed computing on the Semantic Web requires fast execution times. When multiple levels of computing are represented in the URI address space there does not exist a direct physical memory correlate and as such, computing is slower.

With respect to the adoption of the RVM model, the model requires RDF data providers to support RVM ``farms" (or open computing spaces) for foreign processes to use for their execution. Furthermore, the model relies on programming languages that are engineered for a pure URI address space and thus, requires the development of algorithms in these languages.

Initial solutions to the aforementioned issues have been previously discussed in \cite{rodriguez:rvm2008}. The next section provides a brief overview of the various RDF programming languages that exist.

\section{RDF Programming Languages}

An RDF programming language is a language that is not only compiled into RDF, but also aimed at directly manipulating an RDF graph. RDF programming languages include:
\begin{itemize}
	\item FABL: an object-oriented RDF programming language \cite{fabl:bureau2001}
	\item Adenosine: an RDF programming language
	\item Ripple: a functional, stack-based RDF programming language \cite{ripple:shinavier2007}
	\item Neno/Fhat: an object-oriented RDF programming language with RDF-encoded virtual machine. \cite{rodriguez:gpsemnet2007}
\end{itemize}
These languages were designed to take explicit advantage of RDF as a data model. Unlike RDF APIs in other languages such as Java, C, etc. these languages do not require the developer to work with two different data models. That is, RDF APIs in other languages require the developer to write in the constructs of the programming environment as well as in the constructs of RDF. With RDF programming languages, the constructs of the language are simply URIs and RDF triples. There is no disjoint experience for the developer \cite{fabl:bureau2001}.

RDF programming languages compile down to RDF and thus, can be accessed like any piece of data on the Semantic Web. Furthermore, unique situations emerge when RDF code is represented across different physical server machines. Because all RDF software is in the same URI address space, there is nothing that prevents the software, much like the data, to by physically distributed. With an RDF virtual machine executing compiled RDF code, it is possible for the virtual machine and the compiled code to be relocated by simply downloading the RDF subgraph to another environment. Thus, instead of migrating large amounts of data to a local environment for processing, the RDF virtual machine and code can be migrated to the remote environment. In this way, the process is moved to the data, not the data to the process.

\section{Conclusion}

The Semantic Web provides an infrastructure that supports an instantiation of a distributed graph of web resources. However, it lacks a distributed infrastructure for processing that graph. For the World Wide Web, the solution to the issue of processing the vast amount of information has been to literally download the entire Web and index and process it at a single local environment. While the content on the World Wide Web is distributed, the means by which the information on the Web is analyzed is not. The Semantic Web need not fall into this same model. With the nearly limitless ways in which a directed labeled graph (an RDF graph) can be searched and manipulated, it would be a disappointment if the information on the Semantic Web was left to centralized repositories to store, index, and provide query functionality.

\section*{Acknowledgements}

Joshua Shinavier read draft versions of this essay and after doing so, provided many useful comments.

\end{document}